\documentclass[10pt]{wlscirep}  % del _fleqn,
\usepackage[utf8]{inputenc}
\usepackage[T1]{fontenc}
\usepackage{multirow}

\title{Fast-Powerformer: A Memory-Efficient Transformer for Accurate Mid-Term Wind Power Forecasting}

\author[1]{Mingyi Zhu}
\author[1]{Zhaoxing Li}

\author[1,*,+]{Qiao Lin}
\author[1,*,+]{Li Ding}

\affil[1]{
Department of Artificial Intelligence and Automation, School of Electrical
Engineering and Automation, Wuhan University, Wuhan 430072, China}

\affil[*]{Corresponding authors. Emails: liding@whu.edu.cn, linqiao@whu.edu.cn}

\affil[+]{these authors contributed equally to this work}


\begin{abstract}
Wind power forecasting (WPF), as a significant research topic within renewable energy, plays a crucial role in enhancing the security, stability, and economic operation of power grids. However, due to the high stochasticity of meteorological factors (e.g., wind speed) and significant fluctuations in wind power output, mid-term wind power forecasting faces a dual challenge of maintaining high accuracy and computational efficiency. To address these issues, this paper proposes an efficient and lightweight mid-term wind power forecasting model, termed Fast-Powerformer. The proposed model is built upon the Reformer architecture, incorporating structural enhancements such as a lightweight Long Short-Term Memory (LSTM) embedding module, an input transposition mechanism, and a Frequency Enhanced Channel Attention Mechanism (FECAM). These improvements enable the model to strengthen temporal feature extraction, optimize dependency modeling across variables, significantly reduce computational complexity, and enhance sensitivity to periodic patterns and dominant frequency components. Experimental results conducted on multiple real-world wind farm datasets demonstrate that the proposed Fast-Powerformer achieves superior prediction accuracy and operational efficiency compared to mainstream forecasting approaches. Furthermore, the model exhibits fast inference speed and low memory consumption, highlighting its considerable practical value for real-world deployment scenarios.


\end{abstract}

\begin{document}

\flushbottom
\maketitle
\thispagestyle{empty}

\section*{Introduction}

In recent years, with the large-scale integration of renewable energy sources, wind power has emerged as a pivotal component in the global transition towards a cleaner and more sustainable energy system. As a renewable energy source that is clean, efficient, and free of carbon emissions during operation, wind power has experienced rapid growth in both installed capacity and generation scale\cite{LIU2019392}. It now constitutes a substantial share of the electricity supply in many countries and plays a crucial role in reducing greenhouse gas emissions and mitigating climate change.
Despite its advantages, wind power is inherently intermittent and highly dependent on meteorological conditions such as wind speed and direction. This results in significant variability and uncertainty in power output, which poses challenges for grid stability and efficient energy management. The non-dispatchable nature of wind energy complicates its large-scale integration into power systems, often requiring additional balancing resources and operational flexibility to accommodate fluctuations\cite{WANG2014304,LIN2020105835}.Therefore, constructing highly accurate and generalizable wind power forecasting models is of great practical significance for improving wind dispatchability, ensuring grid stability, and reducing
wind curtailment

Wind Power Forecasting (WPF) refers to the process of modeling and predicting future wind power output using historical generation records and meteorological data. Based on the prediction time horizon, WPF can be generally divided into short-term forecasting (within several hours), mid-term forecasting (1 to 3 days), and long-term forecasting (weeks or longer).
In modern electricity markets, both spot transactions and medium- to long-term power contracts depend heavily on the accuracy of power generation forecasts. For power traders, accurate WPF serves two essential purposes: first, it enables the formulation of rational bidding strategies and helps avoid penalties resulting from mismatches between declared and actual generation outputs, such as balancing cost allocations; second, it improves contract planning, reducing the risk of energy shortages or surpluses caused by forecast errors\cite{HAQUE20124563}.
With its low-carbon nature and economic advantages, wind power has become increasingly important in power grid operations and electricity trading. However, its inherent variability introduces a "double-edged sword" effect on grid economics. In the short term, wind fluctuations may increase system balancing costs; in the long term, continued advancements in forecasting technologies and market mechanisms are expected to enhance the overall profitability of wind integration\cite{WANG2016960}.
To effectively balance volatility and profitability, accurate wind power forecasting and intelligent dispatching strategies have become indispensable tools for modern grid operations. In particular, mid-term WPF plays a critical role in generation scheduling, market participation, and coordinated load management.
However, compared to short-term WPF, mid-term forecasting is more challenging due to the extended prediction horizon. It requires models capable of capturing long-range dependencies across historical sequences and learning complex nonlinear relationships and periodic patterns among multiple meteorological variables such as wind speed, direction, and temperature\cite{abdelkader2025optimizing}.

In response to the inherent variability and complexity of wind power output, a wide range of Wind Power Forecasting (WPF) approaches have been developed. These methods can be broadly classified into four categories: physical models, statistical models, machine learning models, and deep learning models. Physical models typically rely on wind dynamics and meteorological principles. A representative method is Numerical Weather Prediction (NWP), which simulates future wind speed by solving partial differential equations describing atmospheric motion and maps them to turbine power output using power curves. Although physical models offer strong interpretability, they are highly sensitive to input conditions and computationally expensive, making them less suitable for real-time short- and mid-term applications. Statistical models such as ARIMA and SARIMA utilize autocorrelation in time series to fit trends but struggle to model the nonlinear variability inherent in wind data due to their linear assumptions. Machine learning approaches such as Support Vector Regression (SVR) and Random Forests improve modeling flexibility via nonlinear mappings but often overlook temporal dependencies, treating forecasting as a static regression problem and failing to capture long-range relationships.

Recently, sequence modeling approaches based on deep learning have shown remarkable progress. RNN-based architectures like LSTM and GRU introduce memory capabilities for handling sequential data but suffer from memory decay and limited parallelism when processing long input sequences.
With advances in computational resources, Transformer-based architectures have become a research hotspot in time series modeling due to their parallel computation and global dependency modeling advantages. Variants such as Informer, Reformer, and Autoformer have achieved promising trade-offs between prediction accuracy and efficiency.

While Transformer-based models have achieved remarkable progress in time series forecasting tasks, driven by their capability to model long-range dependencies and enable parallel computation, their original design stems largely from natural language processing, which differs significantly from the structural characteristics of wind power data. As a result, certain limitations emerge when adapting these models directly to wind power forecasting applications.
In multivariate wind forecasting, for instance, meteorological variables often exhibit strong interdependencies. However, most existing architectures adopt shared attention mechanisms without explicitly modeling these cross-variable interactions. Furthermore, wind power time series are characterized not only by long-term trends but also by localized fluctuations and periodic patterns (e.g., diurnal or seasonal cycles), which are often underrepresented in global attention-based modeling approaches.
Moreover, mid-term forecasting tasks typically require longer input horizons and wider prediction windows compared to short-term tasks, significantly amplifying the computational burden of Transformer models. This becomes particularly problematic in resource-constrained deployment environments such as wind farms or regional control centers, where fast inference and low memory usage are critical requirements.

To meet the dual goals of accurate mid-term wind power forecasting and efficient deployment, the paper proposes Fast-Powerformer, a Reformer-based architecture tailored for long-range, multivariate time series modeling. The model integrates three complementary components into a unified framework: a lightweight LSTM embedding to retain short-term dynamics, an input transposition mechanism to enhance cross-variable interactions and reduce effective sequence length, and a Frequency-Enhanced Channel Attention Module (FECAM) to extract periodic patterns via frequency-domain modulation. These components are jointly designed to improve predictive performance while minimizing computational and memory overhead.
Extensive experiments on real-world wind farm datasets across diverse geographic and climatic conditions demonstrate that Fast-Powerformer outperforms existing state-of-the-art models in forecasting accuracy, inference speed, and generalization ability, making it a strong candidate for practical deployment.

\section*{Related Work}

Wind power forecasting has evolved from physics-based modeling methods to data-driven approaches, including statistical learning, machine learning, and deep learning. With the increasing availability of data and computational resources, research focus has shifted from linear modeling to capturing complex nonlinear patterns, cross-variable dependencies, and long-term temporal correlations.

Early research primarily focused on physics-based methods such as Numerical Weather Prediction (NWP), which models future wind speed by solving partial differential equations based on meteorological variables like temperature, pressure, humidity, and surface roughness, and then maps predicted wind speed to power output using turbine power curves\cite{852131}. Although NWP models offer strong interpretability, their heavy reliance on meteorological conditions, high computational cost, and low update frequency limit their practical applicability in short- and mid-term forecasting. In recent years, hybrid methods combining NWP with data-driven approaches have emerged, such as the WDF model, which integrates multi-resolution information to improve forecasting performance across diverse terrains and temporal scales\cite{GE2025122123}.

To reduce the complexity of physical modeling, researchers progressively turned to statistical models. Initial statistical methods were based on autoregressive (AR) and moving average (MA) models, and their combination in ARMA models was used for stationary time series\cite{2013ARIMA}. Further developments led to widely used extensions such as ARIMA, SARIMA, and fractional ARIMA (fARIMA)\cite{2010A}. These models predict trends by linearly combining past observations. However, the stationarity assumption inherent in these models makes them inadequate for handling the nonstationarity, high-frequency fluctuations, and strong nonlinearity commonly present in wind power data\cite{xiang2024short}. Additionally, the limited capacity for modeling variable interactions restricts the accuracy of statistical models in multivariate and multi-step forecasting tasks.

Subsequently, machine learning methods gained popularity as a new approach to modeling nonlinear relationships. Models like Support Vector Machines (SVM)\cite{708428}, Random Forests (RF)\cite{breiman2001random}, and Gradient Boosting Decision Trees (GBDT)\cite{4a848dd1-54e3-3c3c-83c3-04977ded2e71} have been widely used for wind power forecasting due to their ability to capture complex patterns in historical data. For instance, Guo et al. applied Least Squares SVM (LSSVM) to model residuals from statistical models and improve prediction accuracy\cite{GUO20111668}. Nevertheless, most conventional machine learning methods treat forecasting as a static regression task, ignoring the sequential nature of the data and lacking mechanisms for modeling long-term dependencies and local dynamics. Some studies addressed these limitations by integrating signal decomposition (e.g., EMD) with RF in a multi-scale modeling framework\cite{8515901}, and Cai et al. introduced transfer learning into GBDT for quantile regression to enhance probabilistic forecasting under data scarcity, particularly useful for high wind penetration power grids\cite{caiprobabilistic}.

To further overcome the limitations of previous methods, researchers began exploring deep learning. Artificial Neural Networks (ANN) were first applied to wind forecasting by Charitopoulos et al. in 2000\cite{2002Very}. Convolutional Neural Networks (CNN), with strong feature extraction capabilities, have been used for capturing spatiotemporal patterns in wind farms, while Recurrent Neural Networks (RNN) and their variants, such as LSTM and GRU, focus on modeling long temporal dependencies\cite{singh2007wind,wilms2019exploiting, zhang2019wind,DING201954}. LSTM has also been combined with statistical and machine learning methods in unified frameworks incorporating EMD, PCA, RF, and LSTM, resulting in end-to-end pipelines that balance prediction accuracy, computational efficiency, and interpretability\cite{wang2022wind}. CNN-LSTM models leverage the local perception of CNN and the temporal memory of LSTM to perform well in multivariate forecasting\cite{2024Wind}. Some works further integrate probabilistic forecasting paradigms with CNN-LSTM and employ adaptive CRPS-based loss optimization to address uncertainty quantification and efficiency trade-offs\cite{10400927}. Enhanced models like C-LSTM\cite{wang2025enhancing} and FDNet\cite{MO2024133514} further improve joint modeling of short- and long-term patterns. However, these models may still struggle with either local feature extraction or complex inter-variable interactions.

The introduction of the Transformer architecture\cite{vaswani2017attention} in 2017 marked a paradigm shift in time series modeling. Based on self-attention, Transformers allow global dependency modeling across sequence positions and have achieved groundbreaking results in natural language processing\cite{huang2023deep}. However, their quadratic attention complexity \(\mathcal{O}(L^2)\) presents a scalability bottleneck for long input sequences. To address this, several efficient variants have been proposed. Informer\cite{zhou2021informer} reduces redundancy using ProbSparse attention, Reformer\cite{kitaev2020reformer} lowers complexity to \(\mathcal{O}(L \log L)\) via Locality-Sensitive Hashing (LSH) and employs reversible residuals to save memory, while Autoformer\cite{wu2021autoformer} introduces seasonal-trend decomposition and autocorrelation mechanisms for improved periodic modeling. Despite these improvements, Transformer-based models still suffer from high memory and computation costs, limiting their applicability in resource-constrained scenarios.

Current mainstream approaches still struggle to achieve a balance between forecasting accuracy, cross-variable modeling capability, and computational efficiency in mid-term wind power forecasting. To tackle this challenge, the paper proposes an efficient architecture named \textbf{Fast-Powerformer}, which builds upon the Reformer backbone and integrates three structural innovations: lightweight LSTM embedding, an input transposition mechanism (inspired by iTransformer), and a Frequency-Enhanced Channel Attention Module (FECAM). Together, these components enable synergistic modeling of short-term dynamics, periodic patterns, and inter-variable dependencies. Fast-Powerformer achieves high prediction accuracy while significantly reducing computational cost, making it well-suited for deployment in resource-constrained environments.

\section*{Methodology}

\subsection*{Model Architecture}
\textbf{Fast-Powerformer} is built upon the Reformer architecture to meet the specific demands of mid-term wind power forecasting, integrating long-sequence modeling with structural enhancements tailored for wind power time series. To effectively capture the cross-variable dependencies, localized fluctuations, and periodic patterns inherent in such data, the model introduces three lightweight yet mutually reinforcing components as part of a unified architecture:

\begin{enumerate}
    \item A \textbf{Transpose Operation} is introduced, effectively reducing the computational and time complexity of the model while significantly enhancing its ability to model cross-variable interactions.
    \item The \textbf{Frequency-Enhanced Channel Attention Module (FECAM)} is integrated into the Reformer backbone, improving the model's sensitivity and ability to capture periodic features in wind power data.
    \item A lightweight \textbf{LSTM encoder} is added at the input of the Transformer to address the limitations of Transformer in extracting local features and to enhance the model's ability to represent short-term dynamic features.
\end{enumerate}

\begin{figure}[ht]
\centering
\includegraphics[width=.9\linewidth]{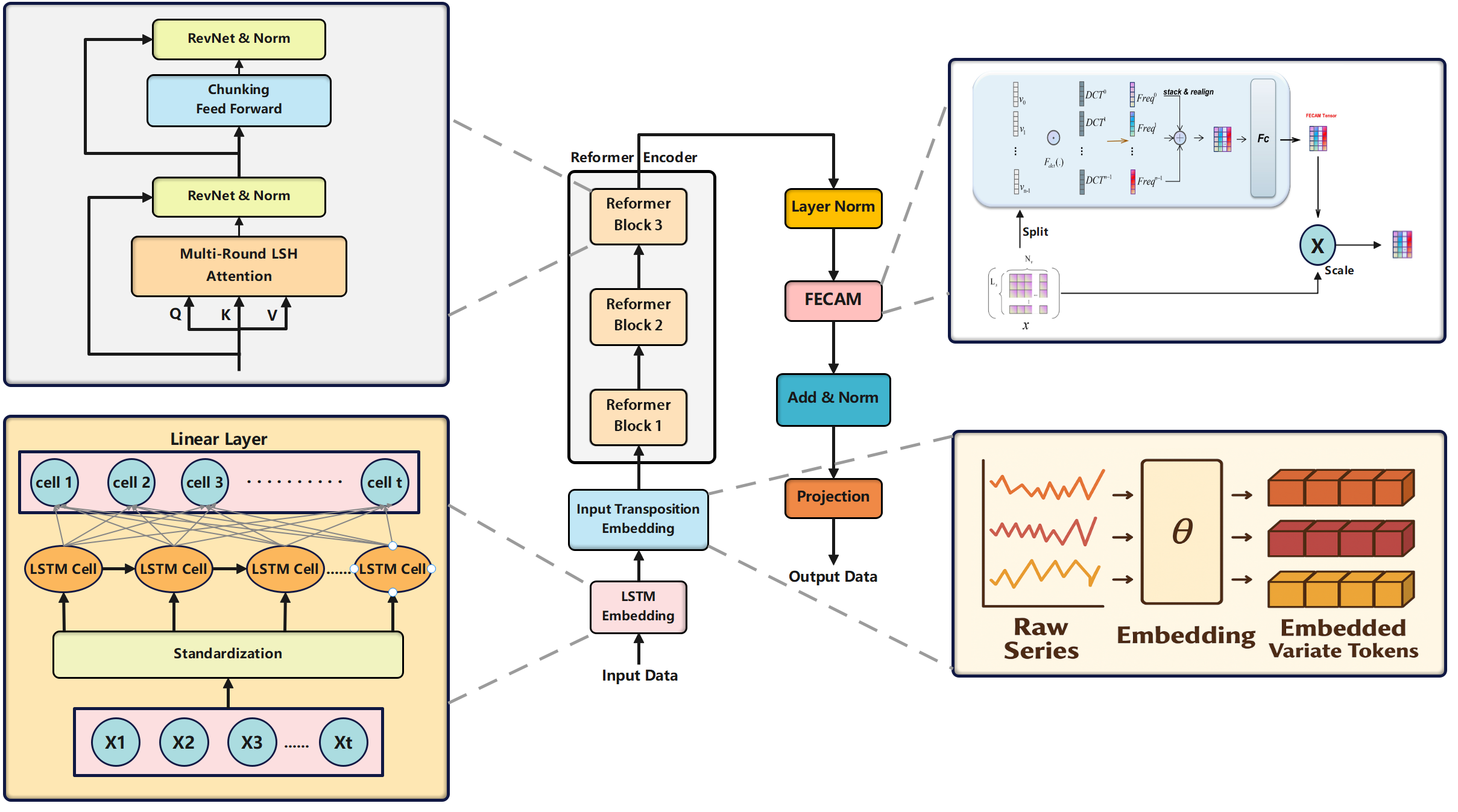}
\caption{Model Architecture}
\label{fig:Architecture Overview}
\end{figure}

The overall framework of the proposed \textbf{Fast-Powerformer} is shown in Figure\ref{fig:Architecture Overview}.

\subsection*{Reformer Backbone}

To efficiently model long-sequence data in mid-term wind power forecasting—particularly under inference and memory constraints—\textbf{Fast-Powerformer} adopts the Reformer architecture. Unlike the standard Transformer, Reformer reduces attention complexity via \textbf{Locality-Sensitive Hashing (LSH)} and lowers memory usage with \textbf{reversible residual layers}.

\subsubsection*{LSH attention mechanism}

The standard self-attention mechanism in Transformer can be expressed as:
\begin{equation}
\text{Attention}(Q, K, V) = \text{softmax} \left( \frac{QK^\top}{\sqrt{d_k}} \right) V
\label{eq:std_attention}
\end{equation}

where \( Q, K, V \in \mathbb{R}^{L \times d_k} \) are the query, key, and value matrices, \( L \) is the sequence length, and \( d_k \) denotes the dimension of each query and key vector. The attention matrix \( QK^\top \in \mathbb{R}^{L \times L} \) results in a computational complexity of \( \mathcal{O}(L^2) \), which becomes prohibitively expensive for long sequences in terms of memory and time cost.

To alleviate this issue, Reformer adopts \textbf{Locality-Sensitive Hashing (LSH)}, which hashes similar queries and keys into the same bucket, reducing the attention computation to within-bucket interactions only. The valid attention scope for a given query position \( i \) is defined as:

\begin{equation}
\mathcal{P}_i = \{ j \mid h(q_i) = h(k_j) \}
\label{eq:lsh_bucket}
\end{equation}
where \( h(\cdot) \) is a locality-sensitive hash (LSH) function that maps high-dimensional vectors into discrete hash buckets. The query vector \( q_i \) and key vectors \( k_j \) are assigned to buckets, and only those keys \( k_j \) sharing the same hash code as \( q_i \) are considered in the attention computation.

The attention output for position \( i \) is computed as:
\begin{equation}
o_i = \sum_{j \in \mathcal{P}_i} \frac{\exp(q_i^\top k_j)}{\sum_{j' \in \mathcal{P}_i} \exp(q_i^\top k_{j'})} v_j
\label{eq:lsh_output}
\end{equation}
where \( \mathcal{P}_i \) denotes the hash-compatible key positions for query \( q_i \), and \( v_j \) is the value vector corresponding to key \( k_j \). The dot product \( q_i^\top k_j \) represents the similarity between the query and key vectors.

To mitigate hash collision errors and improve coverage, Reformer performs multiple independent rounds of hashing. For each query vector \( q_i \), multiple hash functions \( h^{(r)}(\cdot) \), where \( r = 1, \ldots, n_{\text{rounds}} \), are applied. The final attention pool aggregates matches across all rounds:
\begin{equation}
\mathcal{P}_i = \bigcup_{r=1}^{n_{\text{rounds}}} \{ j \mid h^{(r)}(q_i) = h^{(r)}(k_j) \}
\label{eq:multi_round_pool}
\end{equation}
where \( n_{\text{rounds}} \) is the number of hash repetitions. This mechanism reduces the overall attention complexity from the quadratic \( \mathcal{O}(L^2) \) of standard attention to approximately \( \mathcal{O}(L \log L) \), significantly improving scalability for long sequences.

\subsubsection*{Reversible residual layers}

In standard Transformer architectures, each layer must store intermediate activations during the forward pass, leading to memory usage of \( \mathcal{O}(L \cdot d \cdot N) \), where \( L \) is the sequence length, \( d \) is the embedding dimension, and \( N \) is the number of layers. Reformer addresses this issue by introducing a \textbf{Reversible Residual Network}, which allows for the reconstruction of hidden states during backpropagation, thus reducing memory usage.

The reversible residual structure in Reformer splits each layer's hidden state into two partitions, \( X_1 \) and \( X_2 \), and updates them alternately using attention and feed-forward operations. The forward computation is defined as:
\begin{equation}
\left\{
\begin{aligned}
Y_1 &= X_1 + \text{Attention}(X_2) \\
Y_2 &= X_2 + \text{FeedForward}(Y_1)
\end{aligned}
\right.
\label{eq:rev_forward}
\end{equation}
where \( X_1, X_2 \in \mathbb{R}^{L \times d/2} \) represent the two halves of the input hidden state, and \( Y_1, Y_2 \) are the corresponding outputs after applying self-attention and a position-wise feed-forward network, respectively. 

Because the residual update is invertible, the input states can be exactly recovered during backpropagation without storing intermediate activations:
\begin{equation}
\left\{
\begin{aligned}
X_2 &= Y_2 - \text{FeedForward}(Y_1) \\
X_1 &= Y_1 - \text{Attention}(X_2)
\end{aligned}
\right.
\label{eq:rev_backward}
\end{equation}
This reversibility allows the model to reconstruct forward-pass activations from outputs during gradient computation, thereby reducing memory consumption from \( \mathcal{O}(L \cdot d \cdot N) \) to \( \mathcal{O}(L \cdot d) \), where \( L \) is the sequence length, \( d \) is the hidden dimension, and \( N \) is the number of layers.

This structure eliminates the need to store layer-wise activations, thereby reducing memory consumption significantly.

Additionally, Reformer introduces a \textbf{chunked feed-forward} mechanism, which splits the input sequence along the temporal axis and processes chunks independently. This further reduces intermediate memory usage during forward and backward computation.

\subsection*{LSTM Embedding}

Although the Reformer architecture is effective at capturing long-range dependencies in sequential data, its reliance on sparse attention mechanisms can limit its ability to extract local features and short-term dynamics—particularly in high-noise, high-variability scenarios such as wind power forecasting.

\begin{figure}[ht]
\centering
\includegraphics[width=.58\linewidth]{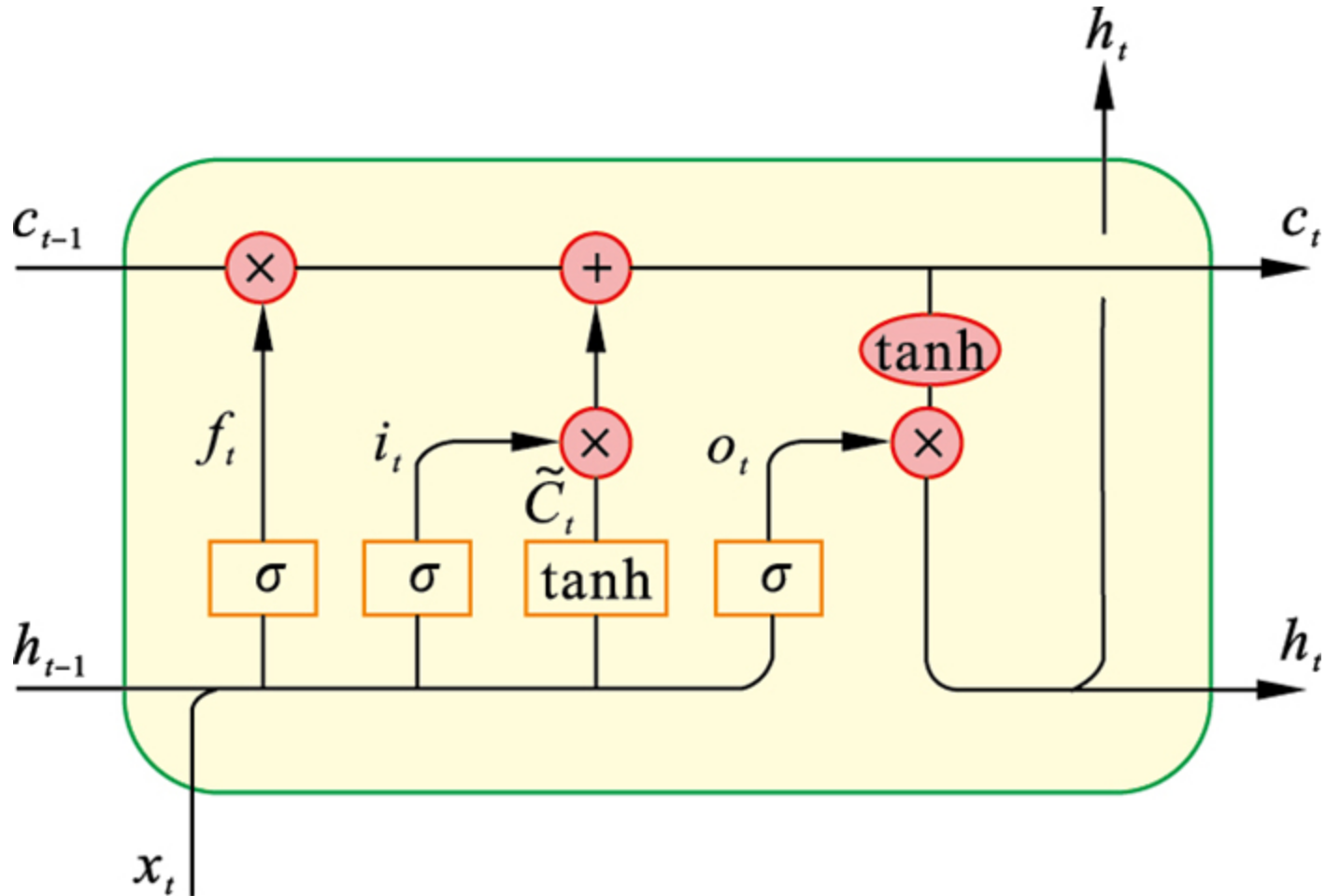}
\caption{Schematic diagram of the internal structure of an LSTM cell.}
\label{fig:lstm}
\end{figure}

To address this limitation, we introduce a lightweight \textbf{Long Short-Term Memory (LSTM)} network as a feature embedding layer preceding the Reformer backbone. LSTM is a specialized variant of recurrent neural networks (RNNs) that incorporates gated mechanisms—namely the input gate, forget gate, and output gate—to effectively model short-term dependencies and temporal patterns in sequences. Unlike standard RNNs, LSTM mitigates the vanishing/exploding gradient problem during long sequence processing, enabling more stable and efficient learning of local dynamics. The internal structure of the LSTM cell is illustrated in Figure~\ref{fig:lstm}.

Given an input sequence \( \mathbf{X} \in \mathbb{R}^{B \times T \times V} \), where \( B \), \( T \), and \( V \) denote the batch size, sequence length, and number of variables respectively, the LSTM updates its hidden states via the following gating operations:
\begin{equation}
\begin{aligned}
f_t &= \sigma(W_f [x_t, h_{t-1}] + b_f), \\
i_t &= \sigma(W_i [x_t, h_{t-1}] + b_i), \\
o_t &= \sigma(W_o [x_t, h_{t-1}] + b_o), \\
\tilde{C}_t &= \tanh(W_c [x_t, h_{t-1}] + b_c), \\
C_t &= f_t \odot C_{t-1} + i_t \odot \tilde{C}_t, \\
h_t &= o_t \odot \tanh(C_t),
\end{aligned}
\label{eq:lstm_gates}
\end{equation}
where \( f_t, i_t, o_t \in \mathbb{R}^{d_h} \) represent the forget gate, input gate, and output gate respectively; \( C_t \) is the cell state and \( h_t \) is the hidden state at time step \( t \); \( \odot \) denotes element-wise multiplication, and \( \sigma(\cdot) \), \( \tanh(\cdot) \) are the sigmoid and hyperbolic tangent functions. These mechanisms enable LSTM to effectively capture local temporal dynamics within the sequence.

After LSTM processing, the resulting hidden feature representation is denoted as:
\begin{equation}
\mathbf{H} = \text{LSTM}(\mathbf{X}), \quad \mathbf{H} \in \mathbb{R}^{B \times T \times d_h}
\label{eq:lstm_output}
\end{equation}
where \( \mathbf{X} \in \mathbb{R}^{B \times T \times V} \) is the input multivariate time series, 
\( B \) is the batch size, 
\( T \) is the sequence length, 
\( V \) is the number of input variables, 
and \( d_h \) is the hidden dimension of the LSTM output.

To align the LSTM output with the Reformer backbone, we apply a linear transformation to project \( \mathbf{H} \) into the target model dimension \( d_{\text{model}} \):
\begin{equation}
\mathbf{E} = \mathbf{H} \mathbf{W}_e + \mathbf{b}_e, \quad \mathbf{E} \in \mathbb{R}^{B \times T \times d_{\text{model}}}
\label{eq:linear_projection}
\end{equation}
where \( \mathbf{W}_e \in \mathbb{R}^{d_h \times d_{\text{model}}} \) and \( \mathbf{b}_e \in \mathbb{R}^{d_{\text{model}}} \) are the learnable weight matrix and bias vector of the projection layer, respectively. 
The resulting projected embedding \( \mathbf{E} \) matches the expected input dimensionality of the Reformer encoder.

This embedding design allows the model to effectively capture short-term sequential features, alleviate optimization challenges, and accelerate convergence, thereby enhancing the overall performance of the Fast-Powerformer.

\subsection*{Input Transposition Mechanism}

In conventional Transformer-based architectures, each time step is treated as a token, and self-attention is computed along the temporal dimension to model long-range dependencies. However, this time-step-centric design presents two key limitations in the context of wind power forecasting.

First, wind power output is typically influenced by complex interactions among multiple variables such as wind speed, direction, temperature, and humidity. Modeling only temporal dependencies neglects \textit{cross-variable dependencies}, limiting the model's ability to capture inter-variable dynamics. Second, due to the long historical windows commonly used in mid-term forecasting, treating each time step as a token leads to an excessive number of tokens. This dramatically increases computational complexity and memory consumption, posing serious constraints for real-world deployment.

To address these limitations, the paper adopts the variable-dimension encoding strategy introduced in iTransformer \cite{liu2023itransformer} and proposes an \textbf{Input Transposition Mechanism}. Specifically, we transpose the input tensor \( \mathbf{X} \in \mathbb{R}^{B \times T \times V} \)—where \( B \), \( T \), and \( V \) represent batch size, sequence length, and number of variables, respectively—into a variable-centric representation:
\begin{equation}
\mathbf{X}^\prime = \text{Permute}(\mathbf{X}), \quad \mathbf{X}^\prime \in \mathbb{R}^{B \times V \times T}
\label{eq:transpose}
\end{equation}

In this format, each variable's temporal profile becomes an independent token. These tokens are passed through a linear layer to project the variable dimension to a higher-dimensional feature space \( d_{\text{model}} \). Self-attention is then computed across variables, rather than across time, using:
\begin{equation}
\text{Attention}(Q, K, V) = \text{softmax}\left(\frac{QK^\top}{\sqrt{d_k}}\right) V, \quad Q, K, V \in \mathbb{R}^{B \times V \times d_k}
\label{eq:cross_variable_attn}
\end{equation}

This mechanism allows each token to represent one variable across all time steps, significantly improving the model’s ability to capture cross-variable interactions. Moreover, since the number of variables \( V \) is typically much smaller than the sequence length \( T \) (i.e., \( V \ll T \)), this transposition reduces the token count in the attention mechanism from \( T \) to \( V \), thereby lowering both computational and memory costs.

By restructuring the input representation in this way, the proposed method not only enhances the modeling of inter-variable dependencies but also greatly improves inference efficiency, making it highly suitable for resource-constrained deployment scenarios in industrial wind power forecasting.

\subsection*{Frequency Enhanced Channel Attention Mechanism}

In multivariate time series tasks such as wind power forecasting, the output power is influenced by multiple meteorological variables (e.g., wind speed, direction, temperature) and their interactions. Additionally, wind power data often exhibit strong periodic characteristics. Traditional time-domain models mainly focus on capturing dynamic dependencies across time, while overlooking rich information embedded in the frequency domain—especially low-frequency components that are crucial for revealing long-term periodic patterns.

\begin{figure}[ht]
\centering
\includegraphics[width=.95\linewidth]{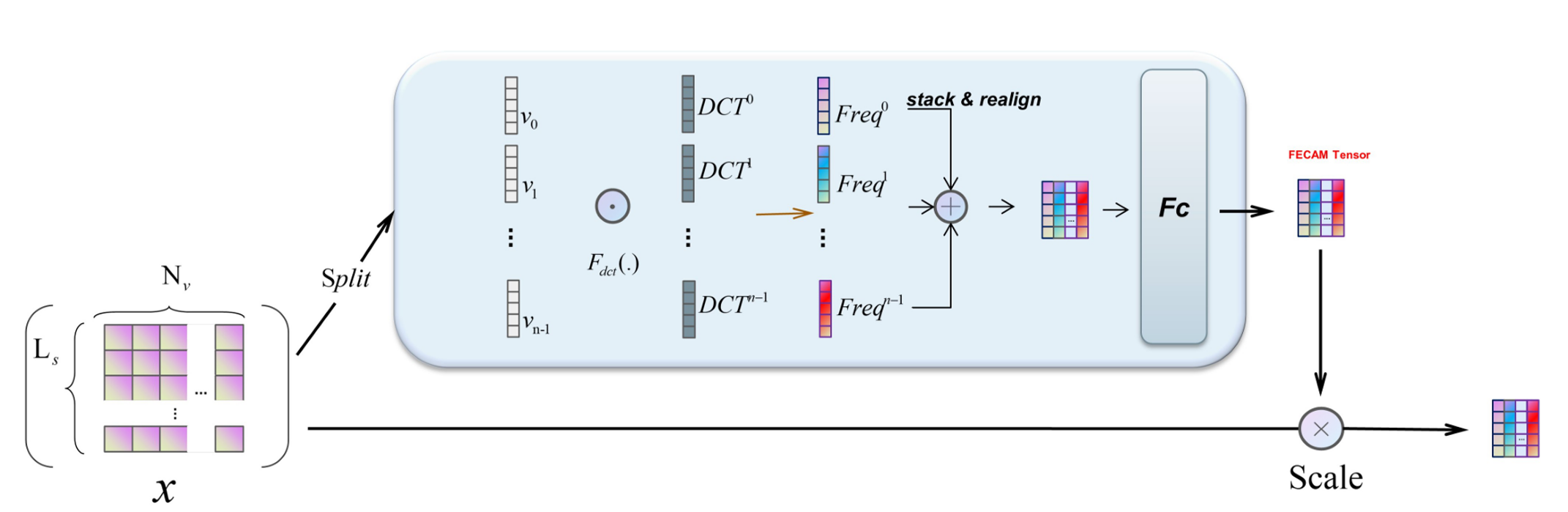}
\caption{Schematic of the FECAM.}
\label{fig:fecam}
\end{figure}

To better capture such periodic features and model complex variable interactions, we integrate a \textbf{Frequency-Enhanced Channel Attention Mechanism (FECAM)} \cite{JIANG2023102158} into the main architecture. FECAM leverages \textbf{Discrete Cosine Transform (DCT)} to extract frequency-domain features and adaptively models inter-channel frequency dependencies. The internal structure of FECAM is illustrated in Figure~\ref{fig:fecam}.

%插图

FECAM first partitions the input feature map along the channel dimension into \( n \) subgroups \( v_0, v_1, \ldots, v_{n-1} \), where each group \( V_i \in \mathbb{R}^{1 \times L} \) corresponds to the time series of a single variable. For each subgroup, DCT is applied to extract frequency components:
\begin{equation}
\mathrm{Freq}_i = \mathrm{DCT}(V_i) = \sum_{l=0}^{L-1} V_i(l) B_l
\label{eq:dct_single}
\end{equation}
where \( V_i \in \mathbb{R}^{1 \times L} \) denotes the time-series feature of the \( i \)-th channel, 
\( V_i(l) \) is its value at position \( l \), and \( B_l \) is the \( l \)-th basis function of the Discrete Cosine Transform (DCT). 
This operation extracts the frequency-domain representation \( \mathrm{Freq}_i \in \mathbb{R}^{1 \times L} \) for each channel.

The frequency-transformed vectors from all subgroups are stacked to form a complete frequency representation:
\begin{equation}
\mathrm{Freq} = \mathrm{DCT}(V) = \mathrm{stack}\left( [\mathrm{Freq}_0, \mathrm{Freq}_1, \ldots, \mathrm{Freq}_{n-1}] \right)
\label{eq:dct_stack}
\end{equation}
where \( n \) is the total number of channels (variables), and \( \mathrm{Freq} \in \mathbb{R}^{B \times V \times L} \) denotes the full frequency matrix, 
with \( B \) the batch size and \( V \) the number of variables.

To capture the relative importance of frequency components across channels, FECAM uses a two-layer feed-forward network to generate attention weights:
\begin{equation}
F_{\text{c-att}} = \sigma\left(W_2 \, \delta(W_1 \, \mathrm{Freq}) \right)
\label{eq:fecam_att}
\end{equation}
where \( W_1 \in \mathbb{R}^{L \times d_{\text{int}}} \), \( W_2 \in \mathbb{R}^{d_{\text{int}} \times 1} \) are the weights of two fully connected layers, 
\( \delta(\cdot) \) is a nonlinear activation function such as ReLU, and \( \sigma(\cdot) \) is a sigmoid function that scales each channel’s response to [0, 1]. 
The resulting attention map \( F_{\text{c-att}} \in \mathbb{R}^{B \times V \times 1} \) reflects the frequency-aware importance of each channel.

The attention map is then applied to the original input \( x \in \mathbb{R}^{B \times V \times L} \) through element-wise multiplication:
\begin{equation}
\mathrm{Output} = x \cdot F_{\text{c-att}}
\label{eq:fecam_output}
\end{equation}

Finally, residual addition and layer normalization are used to stabilize training and preserve the original semantics:
\begin{equation}
x = \mathrm{LayerNorm}(x + \mathrm{Output})
\label{eq:fecam_norm}
\end{equation}

Through this design, FECAM is able to more effectively extract periodic patterns in wind power data and enhance the model’s capacity to learn multivariate interactions, significantly improving overall forecasting performance.

\subsection*{Design Rationale and Inference Workflow}

The overall design of \textbf{Fast-Powerformer} follows a modular but coherent modeling strategy aimed at addressing the structural complexities of wind power time series—specifically long-term dependencies, short-term fluctuations, multivariate coupling, and periodic patterns. Each component is positioned to contribute complementary capabilities at different stages of the prediction pipeline.

The \textbf{LSTM embedding} is placed at the very beginning of the model to enrich the raw input with short-term temporal features. Since recurrent mechanisms are particularly effective in capturing localized temporal dynamics, this stage ensures that early representations already encode essential short-term variation, which is often lost in pure attention-based models.

The \textbf{Input Transposition Mechanism} is applied immediately after LSTM encoding. By shifting the modeling focus from temporal to variable-wise representation, this operation enhances cross-variable interaction modeling while simultaneously reducing the effective sequence length in the attention computation—thereby improving both representational capacity and inference efficiency.

The \textbf{FECAM module} is integrated into the output layers of the Reformer backbone, where it enriches the learned hidden states with periodic frequency-domain information. By applying frequency-aware channel attention, this module helps the model distinguish meaningful periodic patterns from random fluctuations, leading to improved forecasting accuracy and robustness.

To summarize the full inference pipeline, the following pseudocode in Algorithm 1 outlines the step-by-step process of Fast-Powerformer.

\begin{table}[ht]
\centering
\renewcommand{\arraystretch}{1.2}
%\caption*{\textbf{Algorithm 1:} Inference procedure of the proposed Fast-Powerformer model for wind power forecasting}
\begin{tabular}{p{0.95\linewidth}}
\toprule
\textbf{Algorithm 1:} Inference procedure of the proposed Fast-Powerformer model for wind power forecasting
\\ \midrule

\textbf{Input:} Raw multivariate time series $\mathbf{X} \in \mathbb{R}^{B \times T \times V}$ \\ 
\textbf{Output:} Forecasted wind power values $\hat{\mathbf{Y}} \in \mathbb{R}^{B \times P}$ \\
\textbf{1. Preprocessing and temporal encoding:} \\
\quad Apply standardization to $\mathbf{X}$ \\
\quad Encode short-term dynamics with LSTM across time: \\
\quad $\mathbf{H} \leftarrow \text{LSTM}(\mathbf{X}) \in \mathbb{R}^{B \times T \times d_h}$ \\
\quad Project to model dimension: $\mathbf{E} \leftarrow \text{Linear}(\mathbf{H})$ \\
\textbf{2. Input transposition and variate token embedding:} \\
\quad Transpose time and variable dimensions: \\
\quad $\mathbf{E}' \leftarrow \text{Permute}(\mathbf{E}) \in \mathbb{R}^{B \times V \times T}$ \\
\quad Tokenize each variable sequence into a single vector \\
\textbf{3. Long-range modeling via Reformer blocks:} \\
\quad Pass token sequence through $N$ Reformer blocks: \\
\quad $\mathbf{Z} \leftarrow \text{ReformerEncoder}(\mathbf{E}')$ \\
\quad Each block: Multi-round LSH Attention $\rightarrow$ RevNet $\rightarrow$ Chunked FFN \\
\textbf{4. Periodicity modeling via FECAM:} \\
\quad Apply LayerNorm: $\mathbf{Z}_{\text{norm}} \leftarrow \text{LayerNorm}(\mathbf{Z})$ \\
\quad Extract frequency-aware weights via DCT: \\
\quad $\mathbf{A}_{\text{freq}} \leftarrow \text{FECAM}(\mathbf{Z}_{\text{norm}})$ \\
\quad Rescale: $\mathbf{Z}' = \mathbf{Z}_{\text{norm}} \odot \mathbf{A}_{\text{freq}}$ \\
\quad Residual add and normalization: \\
\quad $\mathbf{Z}_{\text{final}} \leftarrow \text{LayerNorm}(\mathbf{Z}' + \mathbf{Z})$ \\
\textbf{5. Output generation:} \\
\quad Project to target dimension and predict: \\
\quad $\hat{\mathbf{Y}} \leftarrow \text{MLP}(\text{Projection}(\mathbf{Z}_{\text{final}}))$ \\
\textbf{Return:} Predicted wind power values $\hat{\mathbf{Y}}$ \\
\bottomrule
\end{tabular}
\label{tab:fast-powerformer-pseudocode}
\end{table}

\section*{Experiments}

\subsection*{Datasets}

The wind power datasets used in the paper are obtained from publicly available real-world data released by the State Grid Corporation of China\cite{chen2022solar}. These datasets contain historical power output and meteorological observations from multiple wind farms operating under real conditions, offering high temporal resolution and strong engineering relevance.
From this collection, the paper selected three representative wind farms, denoted as Farm 1, Farm 2, and Farm 3, to serve as the primary data sources for model training and evaluation. These wind farms are geographically distributed across northern, central, and northwestern China, respectively, covering a diverse range of typical wind power development terrains, including desert, mountainous, and plain regions. They exhibit distinct differences in wind resources and climatic conditions.

All three farms provide data sampled at 15-minute intervals over a continuous two-year period from January 1, 2020, to December 31, 2021, offering sufficient temporal coverage to support mid-term wind power forecasting.Each dataset contains key meteorological variables such as wind speed, wind direction, temperature, atmospheric pressure, and humidity, recorded at multiple altitude levels, including hub height. These variables constitute the essential features required for accurate wind power forecasting. Detailed descriptions of the input variables for each dataset are summarized in Table~\ref{tab:features}.

\begin{table}[htbp]
	\centering
	\caption{Variables of the wind farm datasets.}
	\label{tab:features}
	\begin{tabular}{@{}cl@{}}
		\toprule
		\textbf{Index} & \textbf{Feature name} \\ \midrule
		1  & Time (year-month-day h:m:s) \\
		2  & Wind speed at height of 10 meters (m/s) \\
		3  & Wind direction at height of 10 meters (\textdegree) \\
		4  & Wind speed at height of 30 meters (m/s) \\
		5  & Wind direction at height of 30 meters (\textdegree) \\
		6  & Wind speed at height of 50 meters (m/s) \\
		7  & Wind direction at height of 50 meters (\textdegree) \\
		8  & Wind speed at the height of wheel hub (m/s) \\
		9  & Wind direction at the height of wheel hub (\textdegree) \\
		10 & Air temperature (\textcelsius) \\
		11 & Atmosphere pressure (hPa) \\
		12 & Relative humidity (\%) \\
		13 & Power (MW) (target) \\ \bottomrule
	\end{tabular}
\end{table}

In addition, to comprehensively analyze the wind resource characteristics and meteorological conditions of the selected sites, Table~\ref{tab:statistics} summarizes key statistical metrics for the three wind farms. These metrics include the mean, minimum, maximum, and standard deviation of wind power output, wind speed and wind direction at hub height, as well as near-surface air temperature and relative humidity.
It can be observed that Farm 1 exhibits moderate values across all indicators, reflecting relatively stable operating conditions. Therefore, it serves as the primary dataset for most experiments. Farm 2 has higher average wind speed and power output than the other two farms, indicating more favorable wind resource conditions. In contrast, Farm 3 shows lower wind speeds but exhibits greater variability in temperature and humidity, suggesting a more complex meteorological environment.

\begin{table}[htbp]
	\centering
	\caption{Statistical summary of key variables for each wind farm}
	\label{tab:statistics}
	\resizebox{\textwidth}{!}{
		\begin{tabular}{llccccc}
			\toprule
			\textbf{Wind farm name} & \textbf{Statistics} & \textbf{Power output (MW)} & \textbf{Wind speed at the height of wheel hub (m/s)} & \textbf{Wind direction at the height of wheel hub (°)} & \textbf{Air temperature at 1.5 m (°C)} & \textbf{Relative humidity at 1.5 m (\%)} \\
			\midrule
			\multirow{4}{*}{Farm site 1}
			& Mean     & 23.4  & 6.4  & 217.0  & 8.5   & 37.6 \\
			& Minimum  & 0.0   & 0.0  & 0.0    & $-24.1$ & 0.0 \\
			& Maximum  & 98.1  & 30.2 & 358.5  & 36.1  & 93.1 \\
			& Std. Dev & 24.1  & 3.9  & 85.4   & 13.4  & 18.9 \\
			\midrule
			\multirow{4}{*}{Farm site 2}
			& Mean     & 72.7  & 7.5  & 206.8  & 8.7   & 33.4 \\
			& Minimum  & 0.0   & 0.0  & 0.0    & $-24.5$ & 0.0 \\
			& Maximum  & 201.2 & 28.8 & 359.8  & 37.6  & 97.6 \\
			& Std. Dev & 55.7  & 5.7  & 87.0   & 13.2  & 7.1 \\
			\midrule
			\multirow{4}{*}{Farm site 3}
			& Mean     & 18.1  & 4.0  & 179.1  & 17.4  & 58.5 \\
			& Minimum  & 0.0   & 0.0  & 0.0    & $-14.3$ & 0.0 \\
			& Maximum  & 94.3  & 36.9 & 360.0  & 36.3  & 94.3 \\
			& Std. Dev & 22.6  & 3.3  & 110.5  & 9.9   & 23.8 \\
			\bottomrule
		\end{tabular}
	}
\end{table}

All raw data were preprocessed prior to model training through missing value imputation, outlier correction, and normalization, in order to ensure the effectiveness and stability of the modeling process.

\subsection*{Baselines and Metrics}

To comprehensively evaluate the performance of the proposed \textbf{Fast-Powerformer} in mid-term wind power forecasting tasks, the paper conducted comparative experiments against several representative baseline models. These baselines span traditional statistical methods, classical deep learning architectures, and various mainstream Transformer-based models, specifically:

\begin{itemize}
    \item \textbf{ARIMA}: A classical autoregressive integrated moving average model widely used for linear time series forecasting, representing statistical approaches.
    \item \textbf{MLP}: The multilayer perceptron, a basic feedforward neural network capable of modeling nonlinear mappings.
    \item \textbf{LSTM}: The long short-term memory network, a recurrent architecture with strong sequence modeling capabilities, commonly used in time series prediction.
    \item \textbf{Transformer}: The standard Transformer model based on global self-attention for capturing sequence dependencies, serving as a deep learning baseline.
    \item \textbf{Informer}: An efficient Transformer variant designed for long-sequence forecasting tasks, which incorporates sparse attention mechanisms.
    \item \textbf{Reformer}: A resource-efficient Transformer variant that reduces memory and computational cost via locality-sensitive hashing (LSH) and reversible residual connections.
    \item \textbf{Fast-Powerformer}: The method proposed in this paper, integrating lightweight LSTM embedding, input transposition, and a frequency-enhanced channel attention mechanism, achieving a balance between accuracy and efficiency.
\end{itemize}

To objectively compare the performance of all models, the paper adopted three commonly used evaluation metrics:

\begin{equation}
\mathrm{MSE} = \frac{1}{N} \sum_{i=1}^{N} (y_i - \hat{y}_i)^2
\end{equation}
\noindent
Mean Squared Error (MSE) reflects the average squared difference between predicted and actual values, capturing overall prediction bias.

\begin{equation}
\mathrm{MAE} = \frac{1}{N} \sum_{i=1}^{N} \left| y_i - \hat{y}_i \right|
\end{equation}
\noindent
Mean Absolute Error (MAE) provides a stable and interpretable measure of prediction deviation.

\begin{equation}
\mathrm{MAPE} = \frac{100\%}{N} \sum_{i=1}^{N} \left| \frac{y_i - \hat{y}_i}{y_i} \right|
\end{equation}
\noindent
Mean Absolute Percentage Error (MAPE) evaluates relative prediction error as a percentage of the ground truth, making it suitable for varying data scales.

Here, \( N \) denotes the total number of samples, while \( y_i \) and \( \hat{y}_i \) represent the actual and predicted values at time step \( i \), respectively.

In addition to accuracy, the paper further assessed each model’s deployment efficiency by measuring two practical metrics: the average training time per epoch and the peak memory usage during inference . These indicators reflect the models’ computational cost and deployment feasibility in real-world applications.

\subsection*{Implementation Settings}

All model training and evaluation experiments were conducted under a unified hardware and software environment to ensure fairness and reproducibility. The experiments were implemented using Python~3.10 and PyTorch~2.1, and executed on a single NVIDIA GeForce RTX~3090 GPU with 24~GB of memory, running on Ubuntu~20.04.

All datasets were processed using a sliding window strategy. Each sample consists of the past 3 days of historical observations as the input sequence and predicts the wind power output for the subsequent 3 days. The training, validation, and test sets were partitioned in a 7:1:2 ratio.

All deep learning models were trained using the Adam optimizer, with mean squared error (MSE) as the loss function. The batch size was set to 32. To improve training stability, an early stopping mechanism was employed based on validation set performance, and the best-performing model was retained. To mitigate the effect of random initialization, each experiment was repeated five times, and the average performance metrics are reported as final results.

Detailed hyperparameter settings for both the baseline models and the proposed \textbf{Fast-Powerformer} are provided in Table~\ref{tab:hyperparams}, ensuring fair comparisons under consistent training conditions.

\begin{table}[htbp]
	\centering
	\caption{Parameter settings for all models}
	\begin{tabular}{ll}
		\toprule
		\textbf{Model} & \textbf{Parameters} \\
		\midrule
		Fast-Powerformer & $d_{\text{model}}=128$, $seq\_len=288$, $pred\_len=288$, \\
		& $batch\_size=32$, $epochs=5$, $optimizer=\text{Adam}$, $loss=\text{MSE}$ \\
		\addlinespace[0.5ex]
		Transformer & $d_{\text{model}}=512$, $seq\_len=288$, $pred\_len=288$, \\
		& $batch\_size=32$, $epochs=5$, $optimizer=\text{Adam}$, $loss=\text{MSE}$ \\
		\addlinespace[0.5ex]
		Reformer & $d_{\text{model}}=512$, $seq\_len=288$, $pred\_len=288$, \\
		& $batch\_size=32$, $epochs=5$, $LSH\_rounds=4$ \\
		\addlinespace[0.5ex]
		Informer & $d_{\text{model}}=512$, $seq\_len=288$, $pred\_len=288$, \\
		& $batch\_size=32$, $epochs=5$, $prob\_sparse=\text{True}$ \\
		\addlinespace[0.5ex]
		LSTM & $hidden\_size=128$, $num\_layers=2$, $seq\_len=288$, \\
		& $pred\_len=288$, $batch\_size=32$, $epochs=10$ \\
		\addlinespace[0.5ex]
		MLP & $layers=[128,64,32]$, $activation=\text{ReLU}$, \\
		& $batch\_size=32$, $epochs=10$ \\
		\addlinespace[0.5ex]
		ARIMA & $p=2$, $d=1$, $q=1$ \ (based on AIC criterion) \\
		\bottomrule
	\end{tabular}
	\label{tab:hyperparams}
\end{table}

\section*{Results and Discussion}

\subsection*{Comparison with Baseline Models}

To systematically evaluate the effectiveness of the proposed \textbf{Fast-Powerformer} model for mid-term wind power forecasting, we conducted a series of experiments on the Farm~1 dataset using multiple representative baseline models. The evaluation metrics include Mean Squared Error (MSE), Mean Absolute Error (MAE), and Mean Absolute Percentage Error (MAPE), with the results summarized in Table~\ref{tab:baseline-results}.

\begin{table}[htbp]
	\centering
	\caption{Prediction accuracy comparison of different models for mid-term wind power forecasting}
	\begin{tabular}{lccccccc}
		\toprule
		\textbf{Metrics} & \textbf{Fast-Powerformer} & \textbf{Transformer} & \textbf{Reformer} & \textbf{Informer} & \textbf{LSTM} & \textbf{MLP} & \textbf{ARIMA} \\
		\midrule
		{MSE}  & \textcolor{red}{\textbf{0.851}} & \textcolor{blue}{\textbf{0.898}} & 0.933 & 0.987 & 0.940 & 1.040 & 1.870 \\
		{MAE}  & \textcolor{red}{\textbf{0.652}} & \textcolor{blue}{\textbf{0.664}} & 0.698 & 0.700 & 0.750 & 0.810 & 1.108 \\
		{MAPE} & \textcolor{red}{\textbf{4.736}} & 5.959 & 6.294 & \textcolor{blue}{\textbf{5.057}} & 5.261 & 5.389 & 6.763 \\
		\bottomrule
	\end{tabular}
	\label{tab:baseline-results}
\end{table}

\begin{figure}[ht]
\centering
\includegraphics[width=.8\linewidth]{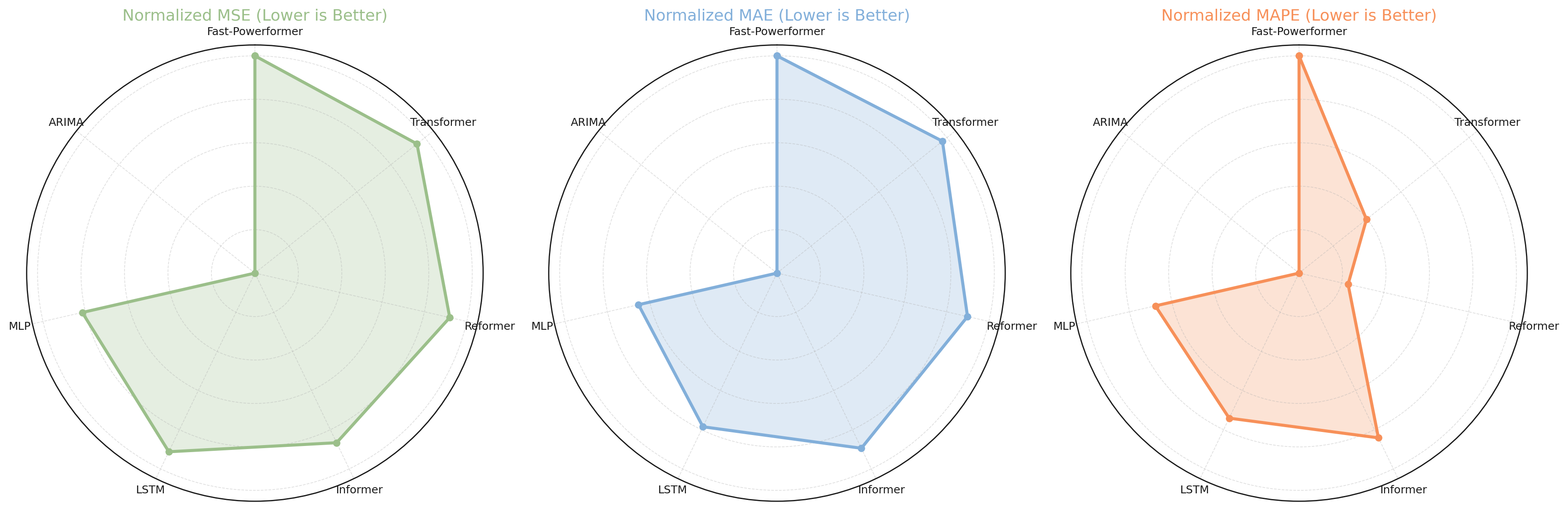}
\caption{Comparison of Forecasting Models via Radar Plot}
\label{fig:radar}
\end{figure}

A radar chart visualization of the model performance is provided in Figure~\ref{fig:radar}. As shown, \textbf{Fast-Powerformer} outperformed all other baselines across all three metrics, achieving an MSE of 0.851, MAE of 0.652, and MAPE of 4.736. Compared to the closest baseline (standard Transformer), Fast-Powerformer reduced MSE and MAE by approximately 5.2\% and 1.8\%, respectively, and achieved a 20.5\% reduction in MAPE. These results highlight the superior modeling capacity of Fast-Powerformer in capturing complex nonlinear temporal patterns.

%插图

In contrast, while Informer achieved a slightly better MAPE (5.057), its MSE (0.987) and MAE (0.700) were noticeably worse, indicating that the ProbSparse attention mechanism might have sacrificed some global trend information. This is particularly evident in scenarios requiring continuous pattern modeling, where local details are critical.

Reformer, although theoretically efficient due to LSH attention and reversible residual layers, showed no substantial performance gain over the standard Transformer, with an MSE of 0.933 and MAPE of 6.294. This suggests limitations in Reformer’s ability to capture cross-variable dependencies and periodic structures in wind power data.

Traditional neural models like LSTM showed further limitations. Its MAE reached 0.750 with MAPE exceeding 5.2, indicating inadequate modeling of long-range dependencies despite its recurrent architecture. MLP performed even worse due to its static nature, failing to leverage temporal dynamics and multivariate interactions.

ARIMA, as a classical linear model, exhibited the poorest performance, highlighting its inadequacy for nonstationary, nonlinear, and multivariate forecasting tasks. Although effective for short-term trend extrapolation, ARIMA struggled to model the complex coupling and nonlinear fluctuations commonly found in wind power data.

\begin{figure}[ht]
\centering
\includegraphics[width=.75\linewidth]{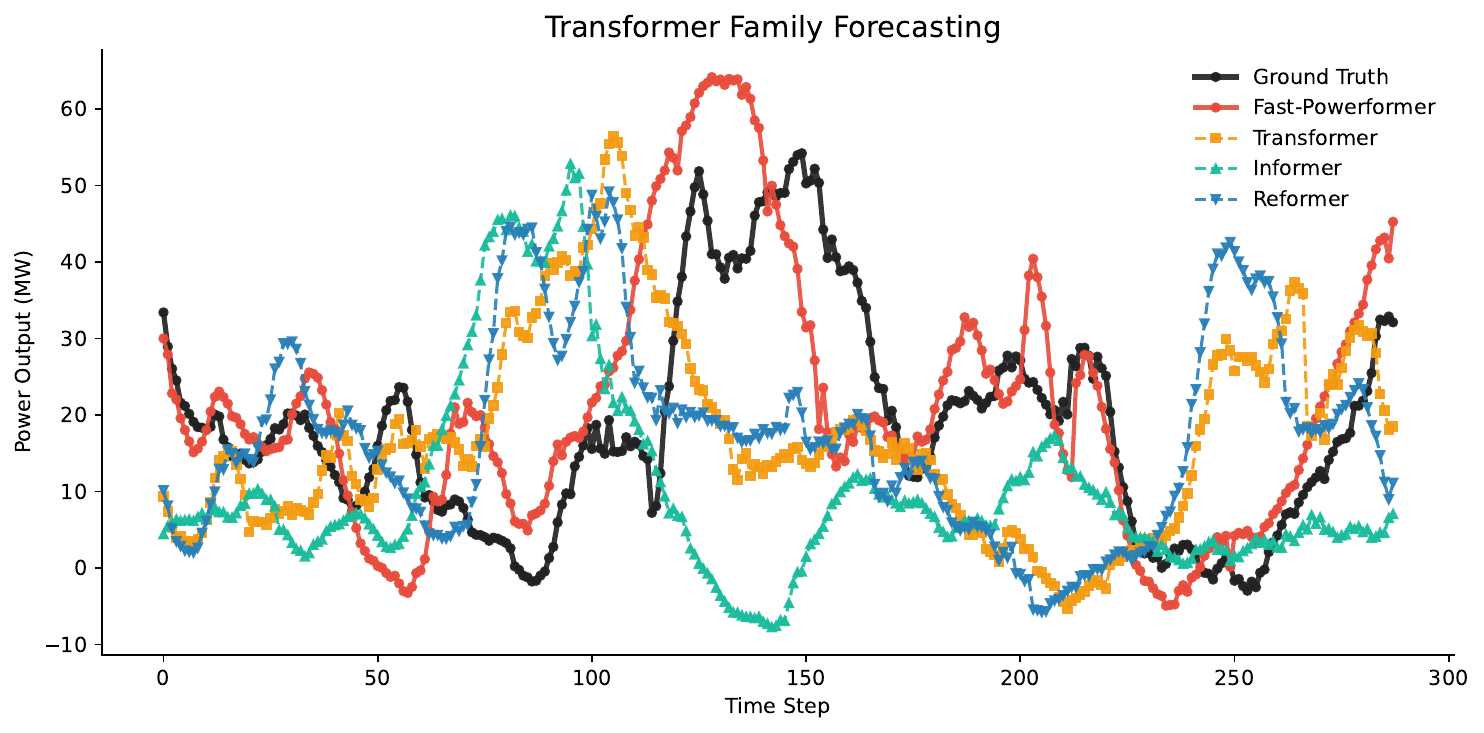}
\caption{Transformer Family Forecasting}
\label{fig:1}
\end{figure}

\begin{figure}[ht]
\centering
\includegraphics[width=.75\linewidth]{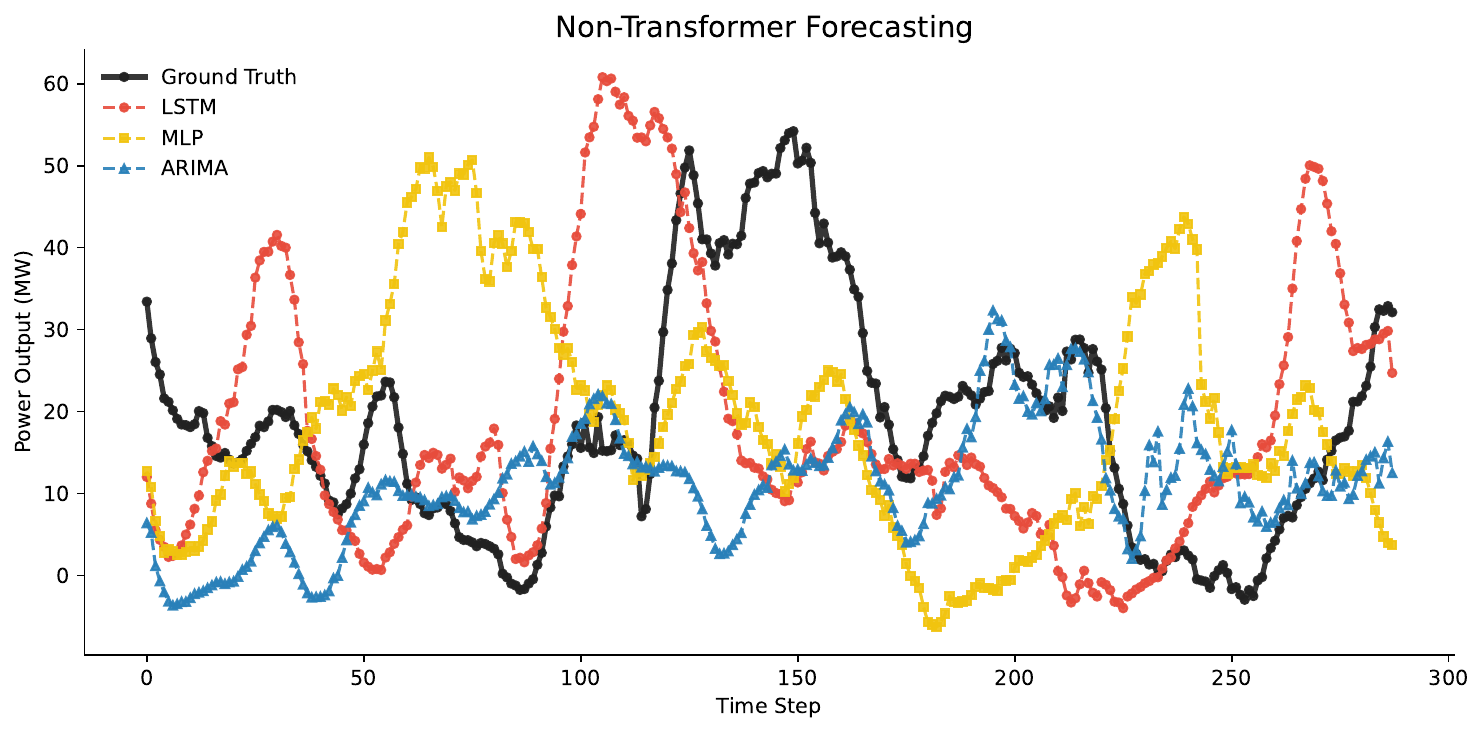}
\caption{Non-Transformer Forecasting}
\label{fig:2}
\end{figure}

Figures~\ref{fig:1} and \ref{fig:2} present a representative case study comparing the prediction curves of different models. \textbf{Fast-Powerformer} demonstrated the closest alignment with the ground truth, not only capturing the global trend but also preserving local fluctuations and turning points with high fidelity.

%插图

%插图

Informer, by contrast, showed noticeable prediction lag and oversmoothing, especially around peaks and valleys, likely due to the sparsity of its attention mechanism. Reformer produced curves similar to Transformer but failed to capture long-range cross-variable dependencies, limiting its ability to model high-frequency volatility and nonstationarity.

Among non-transformer models, LSTM yielded flatter output curves and exhibited delays around sharp changes, indicating its limited effective memory over long sequences. MLP, lacking sequence modeling entirely, produced oversmoothed and oscillatory outputs, failing to capture the true temporal dynamics. ARIMA’s prediction curves deviated sharply from the ground truth, especially in nonlinear segments, reinforcing its unsuitability in this setting.

In summary, \textbf{Fast-Powerformer} achieved not only the best numerical performance but also superior curve fidelity, demonstrating its robustness in both global trend modeling and local detail recovery across complex wind power time series.

\subsection*{Efficiency Analysis}

To further validate the deployment advantages of the proposed \textbf{Fast-Powerformer} in practical applications, the paper conducted a comparative analysis of training efficiency and memory usage across Transformer-based models. Specifically, the paper measured the Epoch Time (training time per epoch) and Memory Usage during training. The results are summarized in Table~\ref{tab:efficiency}.

\begin{table}[ht]
	\centering
	\caption{Inference time and memory usage comparison of Transformer-based models}
	\begin{tabular}{lccc}
		\toprule
		\textbf{Model} & \textbf{Epoch Time (s)} & \textbf{Memory Usage (MB)} & \textbf{Remarks} \\
		\midrule
		Fast-Powerformer & \textcolor{red}{48}   & \textcolor{red}{686}  & Best accuracy, fastest, lowest memory \\
		Informer         & 150                   & 1524                  & Sparse attention helps speed, but less accurate \\
		Reformer         & 200                   & 3682                  & Highest memory, slowest \\
		Transformer      & 125                   & 2558                  & Baseline, but high cost \\
		\bottomrule
	\end{tabular}
	\label{tab:efficiency}
\end{table}

As shown in the results, \textbf{Fast-Powerformer} achieves the best prediction accuracy while demonstrating superior computational efficiency and memory efficiency. It completes a single training epoch in just 48 seconds and uses only 686MB of GPU memory, the lowest among all evaluated models. These advantages can be largely attributed to architectural optimizations: the input transposition mechanism effectively reduces the number of tokens along the temporal dimension, significantly decreasing the computational complexity and memory consumption of attention modules. With lower computational costs and memory requirements, Fast-Powerformer provides a strong foundation for large-scale deployment and real-world engineering applications.
In practical industrial environments such as smart grid dispatching centers or wind farm edge controllers, real-time response, low latency, and limited hardware resources are critical constraints. Fast-Powerformer’s lightweight design directly addresses these challenges.

First, the model exhibits low inference latency and memory footprint, requiring only 686MB of GPU memory and less than 50 seconds per training epoch in our experiments. This ensures compatibility with edge devices or resource-constrained servers, enabling real-time or near-real-time wind forecasting without relying on large cloud clusters.
Second, the architecture is hardware-friendly. It eliminates high-rank attention tensors typically used in full-attention Transformers and avoids complex custom CUDA kernels (e.g., used in FlashAttention). All components, including LSTM, transposition, and FECAM, are implementable using standard deep learning libraries (e.g., PyTorch or TensorRT) with high inference parallelism.
Finally, the model maintains high forecasting accuracy under deployment settings, ensuring reliable power scheduling decisions without the need for post-deployment fine-tuning. These properties collectively make Fast-Powerformer suitable for production-level deployment in real-world energy systems, such as wind power scheduling platforms, renewable energy dispatch systems, or mobile industrial monitoring units.

By comparison, \textbf{Informer}, despite employing a sparse attention mechanism (ProbSparse Attention), achieves faster training time (150s/epoch) than the standard Transformer but suffers from degraded accuracy and memory consumption exceeding 1.5GB. \textbf{Reformer}, while theoretically more efficient due to its locality-sensitive hashing (LSH) attention mechanism, exhibits the highest resource consumption in practice—taking 200 seconds per epoch and consuming 3682MB of memory. This can be attributed to instability in convergence when dealing with high-dimensional multivariate data.

The standard \textbf{Transformer} requires 125 seconds per epoch and 2558MB of memory, confirming its scalability limitations when applied to long-sequence, multivariable forecasting tasks.

\subsection*{Ablation Study}

To assess the individual and joint contributions of key components in \textbf{Fast-Powerformer}, the paper conducts a comprehensive ablation study on the Farm 1 dataset. Built upon the Reformer backbone, the paper evaluates the following modules:

(a) \textbf{Input Transposition Mechanism (ITM)}: restructures input representation to reduce token count and computational complexity;

(b) \textbf{Frequency-Enhanced Channel Attention Mechanism (FECAM)}: captures cross-variable dependencies in the frequency domain;

(c) \textbf{LSTM Embedding}: introduces lightweight temporal modeling to enhance short-term sequence representations.

The experiment covers individual modules, dual combinations , and the full model. Performance and resource consumption metrics for each configuration are reported in Table~\ref{tab:ablation}.

\renewcommand{\arraystretch}{1.2}
\begin{table}[htbp]
	\centering
	\caption{Ablation Study Results of Fast-Powerformer}
	\label{tab:ablation}
	\begin{tabular}{c c c c c c c c c}
		\toprule
		\textbf{ID} & \textbf{Transpose} & \textbf{FECAM} & \textbf{LSTM} & \textbf{MSE} & \textbf{MAE} & \textbf{MAPE} & \textbf{Epoch Time (s)} & \textbf{Memory (MB)} \\
		\midrule
		I   & &             &             & 0.933  & 0.698 & 6.294 &200  & 3682   \\
		II  & \checkmark & \checkmark  &             & 0.853 & 0.659 & 4.931 & 45   & \textbf{404}   \\
		III & \checkmark &             & \checkmark  & 0.911 & 0.703 & 5.121 & \textbf{42}   & 670   \\
		IV  &            & \checkmark  & \checkmark  & \textbf{0.846} & 0.667 & 4.788 & 480  & 4140  \\
		V   & \checkmark & \checkmark  & \checkmark  & {0.851} & \textbf{{0.652}} & \textbf{4.736} & {48} & {686} \\
		\bottomrule
	\end{tabular}
\end{table}

Results indicate that the base Reformer exhibits poor accuracy and high computational cost. Adding ITM and FECAM  significantly improves performance, with marked reductions in MAE and MSE, while also reducing epoch time to 45s and memory usage to 404MB. This demonstrates that the ab combination enhances both efficiency and accuracy.

%插图
\begin{figure}[ht]
\centering
\includegraphics[width=.64\linewidth]{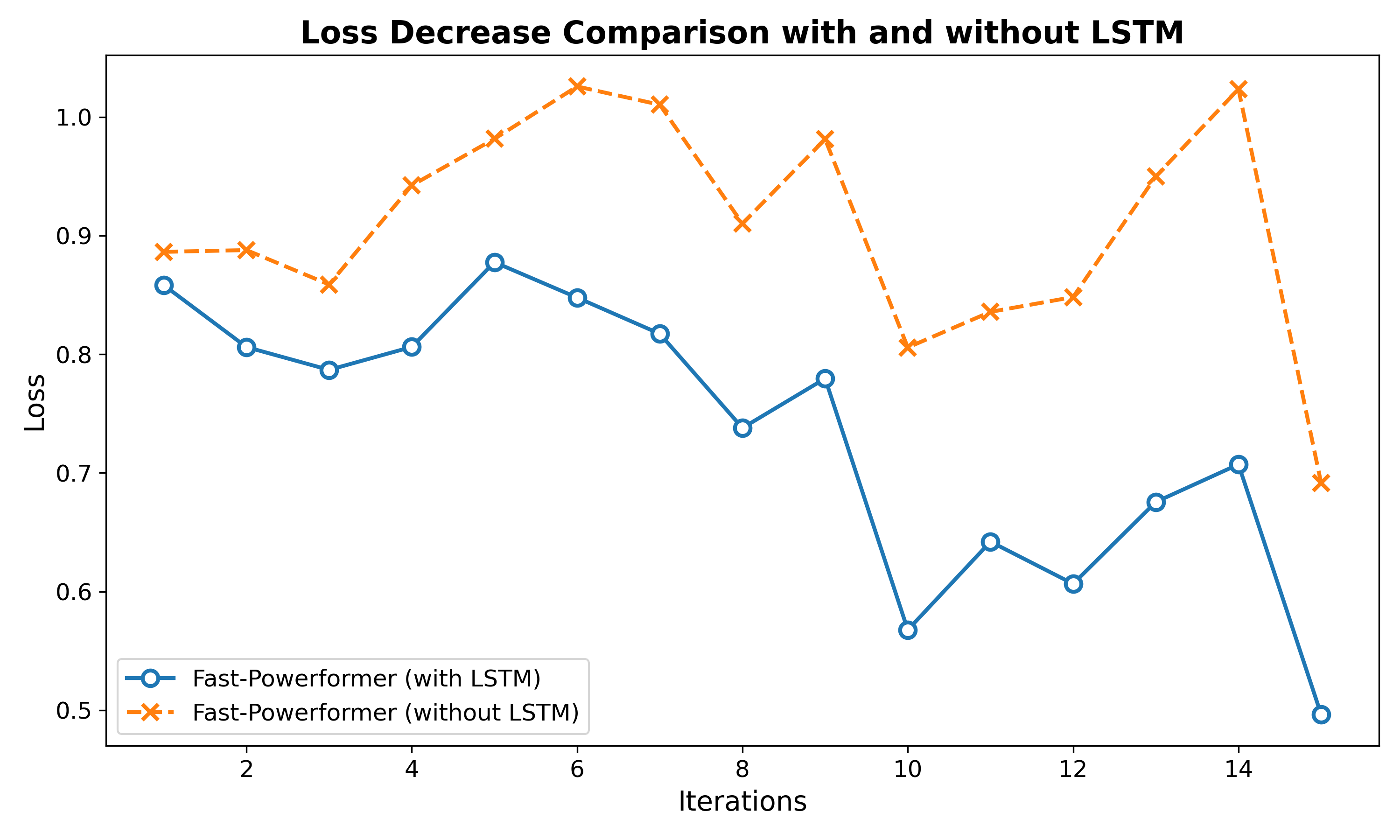}
\caption{Loss Decrease Comparison with and without LSTM }
\label{fig:loss}
\end{figure}

Interestingly, the combination of FECAM and LSTM  yields the best prediction accuracy, highlighting their synergy in capturing temporal and frequency-domain patterns. However, this configuration incurs high resource overhead (480s per epoch, 4140MB memory), exposing Reformer’s limitations with high-dimensional token inputs.

In contrast, the ITM + LSTM configuration  achieves better efficiency (42s, 670MB) but worse accuracy, suggesting that LSTM without frequency guidance (from FECAM) struggles to capture cross-variable dynamics effectively.

Additionally, the paper observes that the inclusion of LSTM substantially accelerates convergence. As illustrated in Figure~\ref{fig:loss}, the LSTM-enhanced Fast-Powerformer exhibits a steeper loss curve compared to its LSTM-free counterpart. Two main factors may explain this:

(1) \textit{Temporal modeling advantage}: LSTM excels at learning short-term dependencies and local temporal dynamics through gated mechanisms, enabling faster convergence.

(2) \textit{Improved training stability}: The LSTM module mitigates gradient vanishing and explosion issues, stabilizing training and accelerating convergence on complex time series data.

\subsection*{Generalization Evaluation}

To evaluate the generalization capability of Fast-Powerformer, the paper further tested the model on two additional wind farm datasets, Farm 2 and Farm 3, which differ significantly in geographic location, terrain type, and climate conditions. These experiments aim to assess the model's adaptability across diverse forecasting scenarios. The results are presented in Table~\ref{tab:generalization}.

\begin{table}[htbp]
	\centering
	\caption{Generalization Evaluation Results on Farm 2 and Farm 3}
	\label{tab:generalization}
	\begin{tabular}{lcccccc}
		\toprule
		\textbf{Model} & \textbf{MSE} & \textbf{MAE} & \textbf{MAPE} & \textbf{Epoch Time (s)} & \textbf{Memory Usage (MB)} \\
		\midrule
		\textbf{Farm 2} & & & & & \\
		\midrule
		Fast-Powerformer & 1.003 & 0.763 & 2.115 & 60 & 688 \\
		Reformer & 1.001 & 0.818 & 2.374 & 302 & 4030 \\
		\midrule
		\textbf{Farm 3} & & & & & \\
		\midrule
		Fast-Powerformer & 0.771 & 0.664 & 3.245 & 55 & 688 \\
		Reformer & 1.016 & 0.785 & 4.191 & 297 & 4386 \\
		\bottomrule
	\end{tabular}
\end{table}

On Farm 2, which represents a region with abundant wind resources, Fast-Powerformer achieved better MAE and MAPE than Reformer, although its MSE was slightly higher. Notably, Fast-Powerformer required only 60 seconds of training time and 688MB of GPU memory, compared to Reformer's 302 seconds and 4030MB, highlighting its efficiency advantage.

On Farm 3, characterized by weaker wind resources and more complex weather patterns, Fast-Powerformer significantly outperformed Reformer across all metrics—achieving 24.1\% lower MSE, 15.4\% lower MAE, and 22.6\% lower MAPE. It also maintained efficient training with only 55 seconds per epoch and 688MB memory usage, compared to 297 seconds and 4386MB for Reformer.

These results demonstrate that Fast-Powerformer consistently delivers superior predictive performance and computational efficiency across varying conditions, confirming its strong generalization ability and suitability for real-world deployment in diverse wind farm scenarios.

\section*{Conclusion}

The paper proposes Fast-Powerformer, an enhanced Transformer-based model tailored for mid-term wind power forecasting. Compared with existing Transformer variants, Fast-Powerformer achieves higher predictive accuracy while significantly improving inference speed and reducing memory usage, making it especially suitable for deployment in resource-constrained environments. To accomplish this, three key architectural innovations are introduced upon the standard Reformer backbone: an Input Transposition Mechanism, a Frequency-Enhanced Channel Attention Module (FECAM), and a lightweight LSTM embedding module. Experimental results demonstrate that Fast-Powerformer excels in capturing periodic patterns in wind power data, enhances short-term dynamic modeling, and offers substantial advantages in training efficiency and memory consumption.

Extensive experiments on three real-world wind farm datasets—Farm 1, Farm 2, and Farm 3—verify the model’s generalization ability across diverse meteorological and wind resource conditions. Fast-Powerformer consistently outperforms classical Transformer-based models (e.g., Reformer, Informer) and deep learning baselines (e.g., LSTM, MLP) in terms of accuracy, efficiency, and computational cost. It exhibits a strong capability in both local pattern recognition and global trend modeling.

Additionally, ablation studies confirm that the input transposition mechanism and FECAM substantially enhance model performance while reducing computational complexity and memory usage. The LSTM embedding module plays a crucial role in accelerating training convergence and improving model stability.

Future work may explore integrating additional multimodal information to further improve forecasting robustness and accuracy, as well as extending the model to long-term forecasting tasks. Reducing computational overhead and memory footprint without compromising accuracy will also remain a key focus of ongoing research.

\section*{Data availability}

All data used in the experiments of this study are derived from the open-source dataset provided by the State Grid Corporation of China. The dataset is publicly available at: \url{https://www.nature.com/articles/s41597-022-01696-6}.

\bibliography{sample}

% For data citations of datasets uploaded to e.g. \emph{figshare}, please use the \verb|howpublished| option in the bib entry to specify the platform and the link, as in the \verb|Hao:gidmaps:2014| example in the sample bibliography file.

%\section*{Acknowledgements (not compulsory)}

% Acknowledgements should be brief, and should not include thanks to anonymous referees and editors, or effusive comments. Grant or contribution numbers may be acknowledged.

\section*{Author contributions}
Zhumingyi was responsible for the model design, all experiments, and the majority of the manuscript writing. Lizhaoxing assisted with parts of the writing and conducted literature review. Linqiao and Dingli provided supervision, guidance, and direction throughout the research and writing process.

\section*{Funding}
This work was supported by the National Natural Science Foundation of
China under Grant 62373290. 

\section*{Declarations}

\section*{Competing interests}
The authors declare no competing interests.

% Must include all authors, identified by initials, for example:
% A.A. conceived the experiment(s),  A.A. and B.A. conducted the experiment(s), C.A. and D.A. analysed the results.  All authors reviewed the manuscript. 

% \section*{Additional information}

% To include, in this order: \textbf{Accession codes} (where applicable); \textbf{Competing interests} (mandatory statement). 

% The corresponding author is responsible for submitting a \href{http://www.nature.com/srep/policies/index.html#competing}{competing interests statement} on behalf of all authors of the paper. This statement must be included in the submitted article file.

% \begin{figure}[ht]
% \centering
% \includegraphics[width=\linewidth]{stream}
% \caption{Legend (350 words max). Example legend text.}
% \label{fig:stream}
% \end{figure}

% \begin{table}[ht]
% \centering
% \begin{tabular}{|l|l|l|}
% \hline
% Condition & n & p \\
% \hline
% A & 5 & 0.1 \\
% \hline
% B & 10 & 0.01 \\
% \hline
% \end{tabular}
% \caption{\label{tab:example}Legend (350 words max). Example legend text.}
% \end{table}

% Figures and tables can be referenced in LaTeX using the ref command, e.g. Figure \ref{fig:stream} and Table \ref{tab:example}.

\end{document}